\documentclass[letterpaper, 10 pt, conference]{ieeeconf}  

\IEEEoverridecommandlockouts                              

\usepackage{amssymb}
\usepackage{amsmath}
\usepackage{amssymb}
\usepackage[noadjust]{cite}
\usepackage{graphicx}
\usepackage{threeparttable}
\usepackage{booktabs}
\usepackage{multirow}
\usepackage{tabularx}
\usepackage{booktabs}
\usepackage{multirow}
\usepackage{dsfont}
\let\labelindent\relax
\usepackage{enumitem}  
\usepackage{float}
\newcolumntype{Y}{>{\centering\arraybackslash}X}
\makeatletter
\let\NAT@parse\undefined
\makeatother
\usepackage[pagebackref=false,breaklinks=true,colorlinks,bookmarks=false]{hyperref}

\hypersetup{
    plainpages=false,
    urlcolor=black,     
    linkcolor=black,    
    citecolor=black,    
    pdfborder={0 0 0}
}

\newif\ifblindreview
\blindreviewfalse  

\title{\LARGE \bf
CubifyGS: Object-Centric 3D Gaussian Splatting for Lifelong Dynamic Scene Maintenance
}

\ifblindreview
    \author{Anonymous Author(s)
    \thanks{Anonymous Institution, Anonymous City, Anonymous Region, Anonymous Country.}%
    \thanks{Correspondence to: Anonymous Author.}%
    \thanks{Acknowledgements and author affiliations are omitted for double-blind review.}%
    \thanks{\rule{0pt}{2\baselineskip}}%
    }
\else
    \author{Bohan Ren$^{1}$, Dianyi Yang$^{1}$, Shiyang Liu$^{1}$, Yu Gao$^{1}$, Jiadong Tang$^{1}$, Zhilin Lai$^{2}$, Yi Yang$^{1}$, Mengyin Fu$^{*, 1}$
    \thanks{This work was partly supported by National Natural Science Foundation of China (Grant No. NSFC 62233002, 92370203).  (*Corresponding Author: M. Fu, fumy@bit.edu.cn)}
    \thanks{$^{1}$School of Automation, Beijing Institute of Technology, Beijing, China}%
    \thanks{$^{2}$Guangzhou Saite Intelligent Technology Co., Ltd.}%
    }
\fi

\begin{document}

\bstctlcite{IEEEexample:BSTcontrol}

\maketitle
\thispagestyle{empty}
\pagestyle{empty}

\begin{abstract}


Lifelong scene mapping under rigid object rearrangement remains a fundamental challenge in robotics. While 3D Gaussian Splatting (3DGS) enables high-fidelity modeling, primitive-level updates often cause persistent ghosting and slow recovery. We propose CubifyGS, an object-level mapping framework that shifts dynamic maintenance from passive re-optimization to active asset management.
CubifyGS models movable instances as reusable Gaussian assets, detects object appearance and disappearance, and updates maps through asset retrieval, rigid transformation, and explicit pruning rather than reconstruction from scratch.
To address geometric voids and local photometric mismatch after such edits, we further propose an event-triggered adaptive optimization strategy that focuses computation on affected regions.
We validate our approach on a newly constructed high-fidelity dynamic benchmark, demonstrating that CubifyGS improves artifact suppression and maintenance efficiency over representative reproducible baselines in the evaluated object-rearrangement setting.
\end{abstract}

\section{INTRODUCTION}

Lifelong scene mapping in dynamic environments is a fundamental challenge in robotics~\cite{cadena2017past}, requiring a robot to accurately model unseen environments and maintain these map representations over time.
Recently, scene mapping techniques based on 3D Gaussian Splatting (3DGS)~\cite{kerbl3Dgaussians} have attracted significant attention owing to their capability for high-fidelity reconstruction of both geometric structures and textures. 
However, most existing 3DGS-based methods~\cite{ha2024rgbd,keetha2024splatam,matsuki2024gaussian,yan2024gs,huang2024photo,zhu2025_loopsplat,wen2025segs,cao2025rgbds,yang2025opengs,yang2025opengs2,liu2025automated} are built upon the assumption of static environments, whereas real-world scenes often exhibit dynamic changes, which frequently manifest as rigid object rearrangements~\cite{wang2023rearrange,ramachandruni2025personalized}.
Consequently, applying these methods directly to such scenarios is fundamentally challenging, as unhandled object movements inevitably corrupt the spatial representation and degrade long-term map consistency.


\begin{figure}
    \vspace{3mm}
    \centering
    \includegraphics[width=1\linewidth]{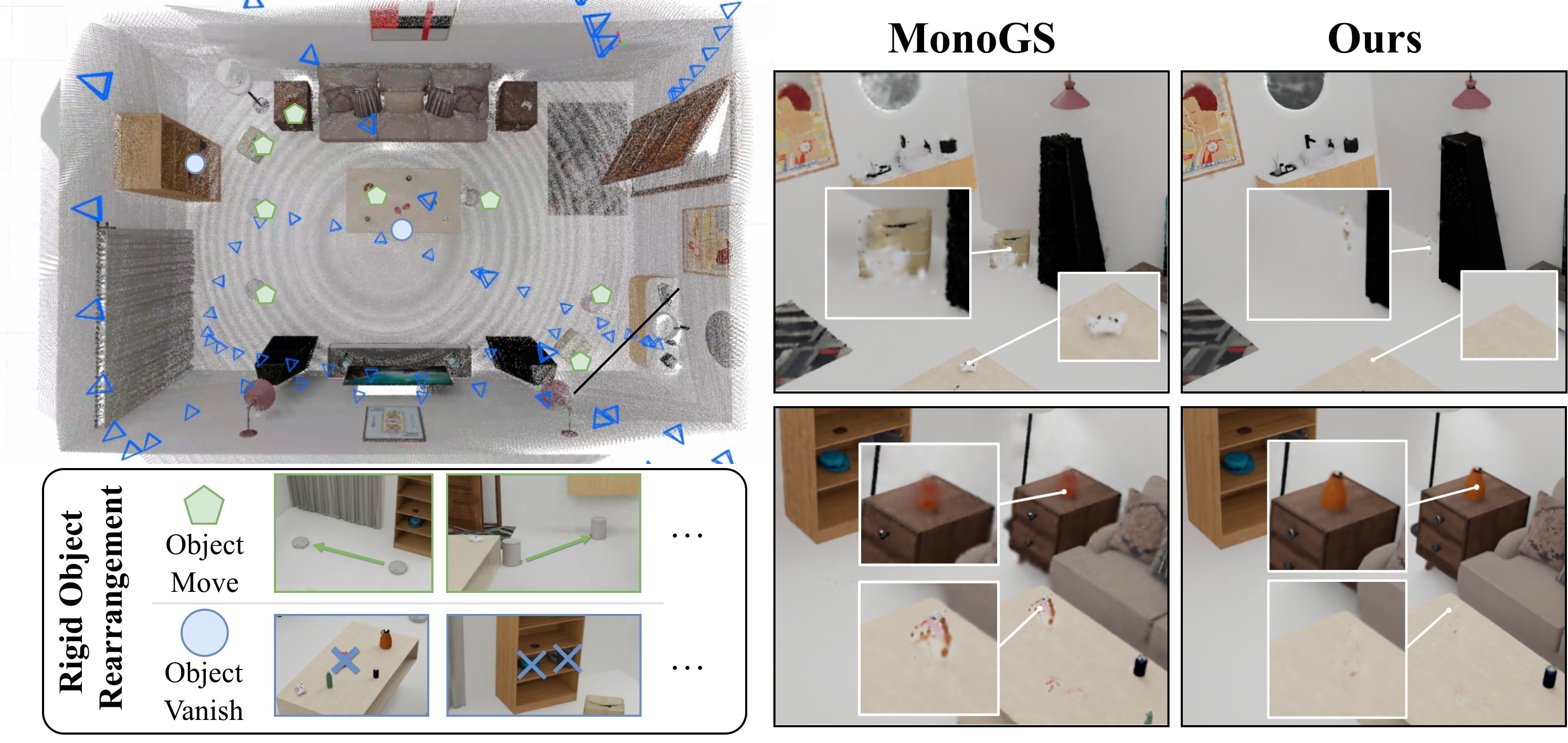}
    \caption{\textbf{Lifelong mapping under rigid object rearrangement.} In dynamic indoor scenes, objects frequently move or vanish (left). Given the same optimization time, conventional gradient-based methods (e.g., MonoGS~\cite{matsuki2024gaussian}) adapt slowly, resulting in noticeable ghosting and incomplete reconstruction (right). In contrast, our object-centric framework explicitly prunes vanished objects and inserts novel ones from an asset library, bypassing slow re-convergence and enabling rapid, high-fidelity, artifact-free scene updates.}
    \label{fig:head_image}
    \vspace{-5mm}
\end{figure}

To address scene dynamics and sustain long-term map consistency, recent solutions~\cite{Zheng2025WildGS,tang2025dronesplat,kong2024dgs,xu2024dg,wen2025gassidy} typically use foundation models or motion cues to filter dynamic objects, or decouple them from the background for separate reconstruction~\cite{li20254d,li2025pg,huang2025adahuman,matsuki20254dtam,zhang2025odhsr,kocabas2024hugs}.
However, they mainly target transient interference or continuous motion, while maintaining 3DGS maps through primitive-level gradient optimization. Under discrete object rearrangement, the system is often forced to implicitly ``erase'' outdated observations and ``reconstruct'' the object at the new location from scratch.
This reactive ``erase-and-reconstruct'' paradigm yields two detrimental consequences: 
(1) \textbf{Persistent Artifacts}: Ghosting remnants at previous locations are difficult to eliminate rapidly, resulting in a pathological map that misleads downstream tasks for extended periods; 
(2) \textbf{Computational Redundancy}: Re-converging the object at its new location from scratch is computationally intensive and slow, often compromising the optimization stability of surrounding static regions~\cite{yang2025bdgs}.



To overcome these limitations, we propose CubifyGS, a lifelong dynamic mapping framework tailored for rigid object rearrangement. Motivated by the insight that indoor dynamics are often object-sparse and reusable~\cite{salas2013slam++,wald2019rio}, we first formulate a \textit{Unified Object-Centric Gaussian Representation} to model movable scene objects and enable streamlined asset management. To reliably perceive environmental changes, we introduce a \textit{Spatio-Temporal Dynamics Perception} module that explicitly detects both the appearance of novel objects and the disappearance of existing ones. Consequently, when encountering object rearrangements, rather than implicitly erasing and re-optimizing geometry via continuous gradients, CubifyGS executes targeted scene maintenance: it explicitly prunes the geometry of vanished objects to instantly eliminate ghosting artifacts, and leverages open-vocabulary grounding~\cite{lazarow2025cubify,lan2025boxfusion,daxberger2025mm,avetisyan2024scenescript,SpatialLM} to retrieve and align reusable templates from a global asset library for newly appeared objects. This shifts maintenance under rigid rearrangement from ``passive re-optimization'' to ``active asset management,'' enabling efficient updates without reconstruction from scratch.


While explicit pruning and retrieval efficiently manage object-level dynamics, they still leave background geometric voids and local photometric mismatches around newly inserted assets. To address these issues, we further introduce an \textit{Event-Triggered Adaptive Optimization Strategy} that increases optimization weights and computation on regions affected by pruning and retrieval, ensuring rapid background inpainting and asset refinement while preserving the stability of the surrounding static environment.

In summary, our contributions are as follows:
\begin{itemize}
    \item We propose CubifyGS, a dynamic 3DGS mapping framework for rigid object rearrangement that represents movable instances as reusable Gaussian assets, shifting maintenance from passive gradient-based re-optimization to active asset management.
    \item We introduce a \textit{Spatio-Temporal Dynamics Perception} module coupled with an \textit{Event-Triggered Adaptive Optimization Strategy}, enabling explicit asset manipulation and rapid local updates without compromising background stability.
    \item We construct a high-fidelity benchmark tailored for dynamic scene mapping with object rearrangement, and experiments demonstrate improved artifact suppression and maintenance efficiency over representative reproducible baselines.
\end{itemize}

\section{RELATED WORK}

\subsection{Object-Centric Proposals and Association for Indoor Scenes}
Recent indoor scene understanding research has shifted towards object-centric representations (e.g., SceneScript~\cite{avetisyan2024scenescript}, SpatialLM~\cite{SpatialLM}, LiteReality~\cite{huang2025literealitygraphicsready3dscene}) that directly export editable 3D bounding boxes. Concurrently, Transformer-based detectors (e.g., CA-1M, CuTR~\cite{lazarow2025cubify}) now provide highly reliable single-frame RGB(-D) 3D box proposals for fine-grained objects. Building upon this, multi-view frameworks like BoxFusion~\cite{lan2025boxfusion} can aggregate per-view proposals into globally consistent, open-vocabulary instances without requiring dense reconstruction. While these advancements offer dependable front-end priors, coupling such object-level entities with long-term 3DGS map maintenance and explicit editing under rigid object rearrangements remains an open challenge.

\subsection{3DGS-based Mapping and Lifelong Maintenance}
Recent dense SLAM systems heavily adopt 3DGS for its high-fidelity rendering. Approaches like SplaTAM~\cite{keetha2024splatam} and MonoGS~\cite{matsuki2024gaussian} utilize photometric optimization for tracking, while GSICP-SLAM~\cite{ha2024rgbd} employs G-ICP for efficient global alignment, and LoopSplat~\cite{zhu2025_loopsplat} leverages high-fidelity 3DGS rendering for robust loop detection and sub-map alignment. More recently, SEGS-SLAM~\cite{wen2025segs} improves photorealistic 3DGS SLAM by introducing structure-enhanced mapping and appearance embeddings. Despite their success on static benchmarks~\cite{straub2019replica,sturm2012benchmark}, these methods do not explicitly model object lifecycle events such as disappearance, reappearance, and relocation, and therefore struggle with rigid object rearrangements that leave persistent ghosting artifacts.

To address dynamics, existing approaches typically mask out moving objects via robust filtering (e.g., DroneSplat~\cite{tang2025dronesplat}, DGS-SLAM~\cite{kong2024dgs}, WildGS-SLAM~\cite{Zheng2025WildGS}, Gassidy~\cite{wen2025gassidy}, Dy3DGS-SLAM~\cite{li2025dy3dgs}) or track continuous non-rigid motion using deformable modeling (e.g., HUGS~\cite{kocabas2024hugs}, 4DGS-SLAM~\cite{li20254d}, 4DTAM~\cite{matsuki20254dtam}). These methods mainly improve robustness to transient dynamic interference or continuously moving objects, whereas our focus is discrete indoor rearrangement where stable objects vanish, reappear, and can be reused as assets. Furthermore, while CL-Splats~\cite{ackermann2025cl} attempts local optimization for changes, its offline nature limits real-time robotic applicability.

Beyond 3DGS, spatio-temporal metric-semantic systems such as Khronos~\cite{schmid2024khronos} and DynaMem~\cite{liu2025dynamem} maintain dynamic scene memories for long-term robot autonomy, object search, and mobile manipulation. They emphasize temporal reasoning with dense metric-semantic maps or point-cloud memories as environments evolve. CubifyGS is complementary to this line of work: rather than constructing a general-purpose spatio-temporal memory, it targets photorealistic 3DGS map maintenance under rigid object rearrangement through explicit ghost pruning, reusable Gaussian asset retrieval, and local adaptive optimization. Bridging these directions remains an interesting avenue for future lifelong robotic mapping.

\section{METHOD}

\begin{figure*}[htbp]
\centering
\includegraphics[width=1.0\linewidth]{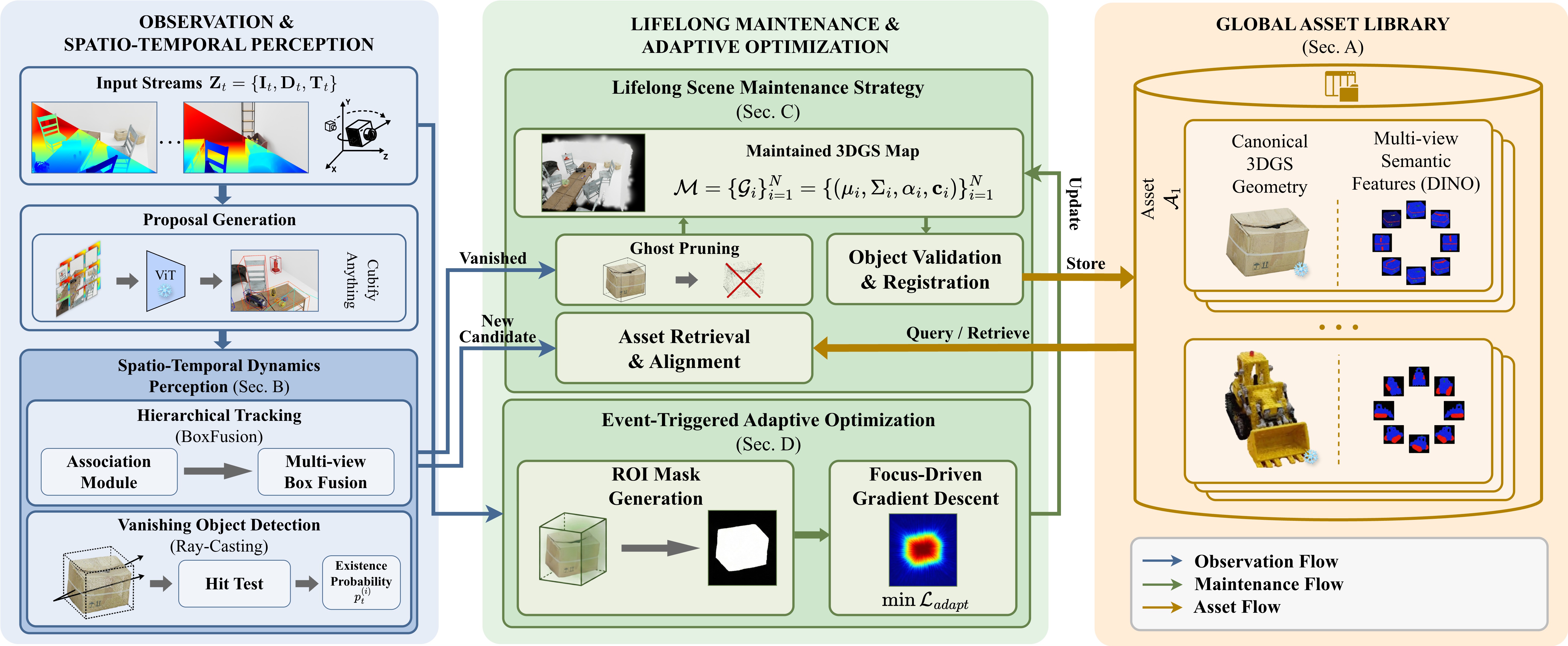} 
\caption{
\textbf{Overview of the CubifyGS framework.} 
Leveraging continuous RGB-D streams, our framework explicitly detects object-level dynamics through hierarchical tracking and ray-casting, and maintains the scene by actively retrieving reusable assets from a global library or pruning ghost artifacts, which are seamlessly integrated via an event-triggered adaptive optimization strategy that concentrates computational resources on dynamic regions for rapid convergence.
}
\label{fig:pipeline_overview}
\vspace{-5mm}
\end{figure*}

We present CubifyGS, an object-centric 3D Gaussian Splatting framework designed for the robust maintenance of indoor maps under rigid object rearrangements. An overview of our pipeline is illustrated in Fig.~\ref{fig:pipeline_overview}. Given a continuous stream of observations $\mathbf{Z}_t = \{ \mathbf{I}_t, \mathbf{D}_t, \mathbf{T}_t \}$, comprising color images $\mathbf{I}_t$, depth maps $\mathbf{D}_t$, and known camera poses $\mathbf{T}_t$, we first model the scene as a unified Object-Centric Gaussian Representation (Sec.~\ref{sec:METHOD_PART_A}). To handle environmental changes, we employ a Spatio-Temporal Perception module to detect object-level discrepancies between current observations and the maintained map (Sec.~\ref{sec:METHOD_PART_B}). Based on these detections, our Lifelong Maintenance Strategy (Sec.~\ref{sec:METHOD_PART_C}) dynamically updates the scene by pruning ghost artifacts and retrieving reusable assets from our Global Asset Library. Finally, to ensure seamless integration and rapid convergence of these updates, we apply an Event-Triggered Adaptive Optimization scheme (Sec.~\ref{sec:METHOD_PART_D}) that focuses computational resources on dynamic regions.

\subsection{Unified Object-Centric Gaussian Representation}
\label{sec:METHOD_PART_A}

Unlike traditional approaches that represent the scene as a monolithic point cloud or a global radiance field~\cite{campos2021orb,davison2007monoslam,mildenhall2021nerf,rosinol2023nerf,sucar2021imap,matsuki2024gaussian,keetha2024splatam,ha2024rgbd}, we formalize the environment as a discretized \textit{object-centric scene representation}, supplemented by a persistent \textit{global asset library}. This dual-representation mechanism enables both high-fidelity rendering and efficient semantic-level manipulation.

\subsubsection{Active Scene List}

We represent the current scene at time $t$ as $\mathcal{S}_t = \{ \mathcal{O}_{bg} \} \cup \{ \mathcal{O}_1, \dots, \mathcal{O}_N \}$, where $\mathcal{O}_{bg}$ denotes the static background and $\{ \mathcal{O}_i \}$ represents the set of dynamic foreground object instances. Each object instance $\mathcal{O}_i$ is encapsulated as a tuple:
\begin{equation}
    \mathcal{O}_i = (\mathcal{G}_i, \mathcal{F}_i, B_i)
\end{equation}

\noindent where the components are defined as follows:

\begin{itemize}[leftmargin=*]
    \item \textbf{Geometry ($\mathcal{G}_i$):} A collection of 3D Gaussian primitives. Each Gaussian is defined by its mean $\mu \in \mathbb{R}^3$, covariance $\Sigma$, opacity $\alpha$, and spherical harmonic (SH) coefficients $\mathbf{c}$~\cite{kerbl3Dgaussians}.
    
    \item \textbf{Semantics ($\mathcal{F}_i$):} To preserve view-dependent characteristics and avoid feature dilution caused by averaging, we maintain a multi-view semantic feature bank $\mathcal{F}_i$. We define a set of $M=12$ fixed canonical viewpoints $\mathbb{V}_{\text{can}} = \{\mathbf{v}^{(1)}, \dots, \mathbf{v}^{(M)}\}$ (e.g., uniformly sampled azimuths). For each object, we store a corresponding set of DINO~\cite{simeoni2025dinov3} features:
    \begin{equation}
        \mathcal{F}_i = \left\{ \mathbf{f}_i^{(k)} = \text{DINO}(I_{\mathbf{v}^{(k)}}) \mid k=1, \dots, M \right\}
    \end{equation}
    This discrete feature bank supports coarse \textit{viewpoint-specific retrieval}, matching current observations with the geometrically most similar viewpoint in the bank.
    
    \item \textbf{Spatial Bounding Box ($B_i$):} A gravity-aligned 3D bounding box that encloses the object's spatial boundaries. It is parameterized by its center, dimensions, and a yaw-only rotation, implicitly defining the object's pose in the world frame.
\end{itemize}

\subsubsection{Global Asset Library}

We maintain a global library $\mathcal{L} = \{ \mathcal{A}_k \}$ storing object templates in a \textit{canonical coordinate system}. Instantiating an asset $\mathcal{A}_k$ as a scene object $\mathcal{O}_i$ involves transforming its Gaussian parameters to world space via:
\begin{equation}
\begin{split}
    \mu_{\text{world}} &= \mathbf{R}_i \mu_{\text{canonical}} + \mathbf{t}_i, \\
    \Sigma_{\text{world}} &= \mathbf{R}_i \Sigma_{\text{canonical}} \mathbf{R}_i^\top
\end{split}
\end{equation}
where the transformation $\mathbf{T}_i = [\mathbf{R}_i | \mathbf{t}_i]$ is derived from the object's bounding box $B_i$. This formulation enables instant asset reuse across different locations without re-optimization.

\subsection{Spatio-Temporal Dynamics Perception}
\label{sec:METHOD_PART_B}

To maintain the \textit{Active Scene List} $\mathcal{S}_t$ consistent with the evolving environment, the system must manage the complete lifecycle of object instances. This involves two coordinated tasks: (1) robustly tracking existing objects and instantiating new candidates from unmatched proposals, and (2) explicitly detecting and pruning objects that have vanished from the scene.

\subsubsection{Hierarchical Tracking and Initialization}
We build our tracking backbone upon the reconstruction-free paradigm BoxFusion~\cite{lan2025boxfusion}, treating object tracking as a probabilistic state estimation problem directly on the image stream. Given the current RGB-D observations $\mathbf{Z}_t$, we first generate a set of 3D object proposals $\mathcal{P}_t$ using a visual foundation model (e.g., CubifyAnything~\cite{lazarow2025cubify}). To ensure persistent identity—crucial for asset retrieval—we employ a hierarchical data association mechanism linking $\mathcal{P}_t$ to existing instances in $\mathcal{S}_{t-1}$:

\begin{itemize}[leftmargin=*]
    \item \textbf{Layer 1 (3D Geometry):} For prominent objects, we compute the volumetric Intersection-over-Union (IoU) between the convex hulls of the proposals and the bounding boxes of existing instances.
    \item \textbf{Layer 2 (2D Projection):} For smaller objects susceptible to depth noise, we project the 3D boundaries of existing objects onto the image plane using the camera pose $\mathbf{T}_t$ and compute the 2D IoU with the visual proposals.
\end{itemize}

A successful match triggers an update of the object's bounding box $B_i$ via a Particle Filter optimizer, maximizing the coherence between the 3D projection and 2D observations. Conversely, unmatched proposals are initialized as candidate new objects, entering the instantiation queue.

\subsubsection{Vanishing Object Detection via Ray-Casting}
Traditional trackers often rely on a static assumption and fail to perceive object disappearance~\cite{lan2025boxfusion}. To address this, we introduce a \textit{Ray-Casted Occupancy Analysis} mechanism to explicitly estimate the \textit{existence probability} $p_t^{(i)}$ of each tracked object.

For each object $\mathcal{O}_i \in \mathcal{S}_t$ within the current view frustum, we define a Region of Interest (ROI) centered at its projection. We cast $N$ rays through the ROI pixels. For each ray $\mathbf{r}_k$, we compute its intersection interval $[t_{\text{in}}, t_{\text{out}}]$ with the object's bounding box $B_i$. By comparing the observed depth $d_k$ with this interval (given a tolerance $\epsilon$), we classify the geometric state $E_k$ of the ray:
\begin{equation}
    E_k = 
    \begin{cases} 
        \textbf{Occluded}, & \text{if } d_k < t_{\text{in}} - \epsilon \\
        \textbf{Hit},      & \text{if } t_{\text{in}} - \epsilon \le d_k \le t_{\text{out}} + \epsilon \\
        \textbf{Free},     & \text{if } d_k > t_{\text{out}} + \epsilon 
    \end{cases}
\end{equation}
The \textbf{Free} state implies the ray penetrated the object's supposed volume without hitting any surface, providing strong evidence of disappearance. We aggregate these ray-level evidences to update the object's existence probability $p_t^{(i)}$ via a temporal decay rule:
\begin{equation}
    p_t^{(i)} = 
    \begin{cases} 
        \min(1, p_{t-1}^{(i)} + \delta_{\text{hit}}), & \text{if } \frac{N_{\text{hit}}}{N} > \tau_{\text{hit}} \\
        p_{t-1}^{(i)} \cdot \lambda_{\text{free}}, & \text{if } \frac{N_{\text{free}}}{N} > \tau_{\text{free}} \land \frac{N_{\text{occ}}}{N} < \tau_{\text{occ}} \\
        p_{t-1}^{(i)} \cdot \lambda_{\text{decay}}, & \text{otherwise}
    \end{cases}
\end{equation}
where $\lambda_{\text{free}} \ll 1$ imposes a severe penalty for confirmed empty space, while $\tau_{\text{occ}}$ prevents false deletion under heavy occlusion. When $p_t^{(i)}$ falls below a threshold $\tau_{\text{vanish}}$ after a grace period, the object is flagged as ``Vanished,'' triggering the pruning and map maintenance process.

\subsection{Lifelong Scene Maintenance Strategy}
\label{sec:METHOD_PART_C}

This module executes the core maintenance operations based on the detections from the previous stage, ensuring the map remains a faithful representation of the current environment.

\subsubsection{Probabilistic Ghost Pruning}
Upon identifying a ``Vanished'' object via the occupancy analysis in Sec.~\ref{sec:METHOD_PART_B}, we trigger the \textit{pruning mechanism}. This explicitly removes all Gaussian primitives confined within the object's bounding box $B_i$. While this effectively eliminates ghost artifacts, it inevitably leaves a geometric void in the background. We defer the rapid filling of these voids to the \textit{Adaptive Optimization} stage (Sec.~\ref{sec:METHOD_PART_D}).

\subsubsection{Semantic-Aware Asset Retrieval \& Alignment}
Upon detecting a newly appearing object, instead of optimizing from scratch, we first query the Global Asset Library $\mathcal{L}$ to retrieve a potentially reusable model.

\textit{Retrieval:} We compute the cosine similarity between the current object's visual feature $\mathbf{f}_{new}$ and the multi-view feature banks of all assets in the library. The best matching asset $\mathcal{A}^*$ and its canonical viewpoint $k^*$ are identified via:
\begin{equation}
    (\mathcal{A}^*, k^*) = \operatorname*{arg\,max}_{\mathcal{A}_j \in \mathcal{L}, k \in \{1, \dots, M\}} \cos(\mathbf{f}_{new}, \mathbf{f}_{\mathcal{A}_j}^{(k)})
\end{equation}
If the maximum similarity exceeds a threshold $\tau_{match}$, we retrieve $\mathcal{A}^*$ for instantiation. The identified canonical viewpoint $k^*$ provides an initial \textit{Coarse Alignment}, aligning the asset's canonical orientation with the current observation.

\textit{Fine Alignment:} To refine the coarse alignment, we leverage the duality between object and camera motion, reformulating object pose estimation as a differentiable camera tracking problem. We freeze the Gaussian parameters and optimize the relative camera pose $\mathbf{T}$ by minimizing the photometric error. Following MonoGS~\cite{matsuki2024gaussian}, we efficiently approximate the gradient by isolating the dominant geometric flow term:
\begin{equation}
    \frac{\partial \mathcal{L}_{pho}}{\partial \mathbf{T}} \approx \sum_{i} \frac{\partial \mathcal{L}_{pho}}{\partial \mathbf{C}} \frac{\partial \mathbf{C}}{\partial \alpha_i} \frac{\partial \alpha_i}{\partial \mathbf{m}_i} \frac{\partial \mathbf{m}_i}{\partial \mathbf{T}}
\end{equation}
where $\mathcal{L}_{pho}$ represents the photometric loss, while $\mathbf{C}$, $\alpha_i$, and $\mathbf{m}_i$ denote the rendered color, opacity, and projected 2D coordinate, respectively. This approximation discards the gradients with respect to covariance and color attributes, focusing solely on the projected means $\mathbf{m}_i$ to drive geometric alignment efficiently.
This formulation avoids the computational overhead of full derivative calculation while ensuring rapid convergence to the precise alignment.

\subsubsection{Candidate Validation and Asset Registration}
To ensure the quality of our Global Asset Library, dynamic objects undergoing optimization are only promoted to the library when they satisfy strict convergence and coverage criteria. We introduce a binary promotion indicator $\mathbb{I}_{promote} \in \{0, 1\}$ defined as:
\begin{equation}
    \mathbb{I}_{promote} = \mathds{1}(C_{view} > \tau_v \land \nabla_{stable} \le \varepsilon)
\end{equation}
where:
\begin{itemize}[leftmargin=*]
    \item $C_{view}$ denotes the \textbf{geometric coverage}, calculated as the angular span of azimuths from which the object center has been observed. This constraint prevents incomplete, ``paper-thin'' reconstructions caused by insufficient multi-view observations.
    \item $\nabla_{stable}$ represents the \textbf{gradient convergence degree}. We consider the optimization stable if the moving average of the gradient magnitudes for the Gaussian attributes remains below the threshold $\varepsilon$ over $K$ consecutive frames, signifying that the object's representation has reached equilibrium.
\end{itemize}

\subsection{Event-Triggered Adaptive Optimization}
\label{sec:METHOD_PART_D}

To address the geometric voids left by pruning and to rapidly adapt newly instantiated (or retrieved) assets to the local photometric environment, we propose an \textit{Event-Triggered Adaptive Optimization} strategy. Unlike traditional methods that treat all pixels uniformly, our approach leverages the locality of indoor changes, concentrating computational resources specifically on the regions affected by dynamic events.

\subsubsection{ROI Mask Generation}
Upon the detection of a change event $\mathcal{E}_i$ (e.g., object addition or removal) at time $t$, we define a 3D Region of Interest corresponding to the object's bounding box $B_i$. For any frame $\tau$ within the sliding window $\tau \in [t, t+w]$, we project this 3D volume onto the 2D image plane to generate a binary interest mask $\mathcal{M}_{roi}^{(\tau)}$:
\begin{equation}
    \mathcal{M}_{roi}^{(\tau)}(\mathbf{u}) = \mathds{1} \left( \mathbf{u} \in \Pi(\mathbf{K}, \mathbf{T}_\tau, B_i) \right)
\end{equation}
where $\mathbf{u}$ denotes the 2D pixel coordinates, $\Pi(\cdot)$ represents the perspective projection operator, and $\mathds{1}(\cdot)$ is the indicator function. This mask explicitly delineates the screen-space region requiring intensive updates.

\subsubsection{Focus-Driven Gradient Descent}
To facilitate rapid scene adaptation, we modulate the standard 3DGS loss function using a spatial weighting map $\boldsymbol{\Omega}$. The adaptive loss $\mathcal{L}_{adapt}$ is formulated as:
\begin{equation}
    \mathcal{L}_{adapt} = \sum_{\mathbf{u}} \boldsymbol{\Omega}(\mathbf{u}) \cdot \left[ (1 - \lambda)\mathcal{L}_1(\mathbf{u}) + \lambda \mathcal{L}_{D-SSIM}(\mathbf{u}) \right]
\end{equation}
where the weighting matrix $\boldsymbol{\Omega}(\mathbf{u})$ assigns a higher importance $\gamma_{focus}$ to the dynamic region while maintaining a baseline weight for the static background to prevent catastrophic forgetting:
\begin{equation}
    \boldsymbol{\Omega}(\mathbf{u}) = 
    \begin{cases} 
        \gamma_{focus}, & \text{if } \mathcal{M}_{roi}(\mathbf{u}) = 1 \\
        1, & \text{otherwise}
    \end{cases}
\end{equation}
By setting $\gamma_{focus} \gg 1$, we force the \textit{gradient flow} to be dominated by the residuals in the changing region. This mechanism ensures that the Gaussian primitives within the ROI (e.g., the newly filled background or the added object) converge rapidly to the new observation, while the rest of the map remains stable.

\section{EXPERIMENTS}
To comprehensively validate CubifyGS, we organize the experiments into the following parts.
In Sec.~\ref{sec:system_comparison}, we present system-level comparisons to evaluate dynamic reconstruction quality and responsiveness under object rearrangement.
In Sec.~\ref{sec:exp_dynamics}, we verify frontend dynamics perception through 3D object localization quality for downstream maintenance.
In Sec.~\ref{sec:exp_retrieval}, we analyze the Retrieval-and-Align pipeline from semantic discriminability, coarse alignment precision, and fine alignment convergence.
Finally, in Sec.~\ref{sec:ablation}, we conduct ablation studies to quantify the contribution of focus-driven optimization and global asset reuse.

\subsection{Experimental Setup}
\label{sec:exp_setup}

\noindent \textbf{Benchmark Construction.}
Existing benchmarks (e.g., TUM RGB-D~\cite{sturm2012benchmark} and Replica~\cite{straub2019replica}) provide limited support for object-level rearrangement, making them less suitable for evaluating lifelong map maintenance.
We therefore build a high-fidelity dynamic indoor benchmark in Blender, covering four scenes (\textit{Bedroom, Office, Living Room, Kitchen}) with discrete rigid-object rearrangement events.
The synthetic pipeline provides pixel-level supervision, including camera trajectories, depth maps, instance masks, and accurate 3D oriented bounding boxes for dynamic objects.
To assess transfer to real data, we additionally evaluate on \texttt{bonn\_kidnapping\_box2}~\cite{palazzolo2019iros}.
Table~\ref{tab:dataset_stats} summarizes sequence statistics and dynamic object counts.

\begin{table}[t]
\vspace{5mm} 
\centering
\caption{Statistics of the Proposed Dynamic Benchmark.}
\label{tab:dataset_stats}
\setlength{\arrayrulewidth}{1.0pt} 
\resizebox{0.48\textwidth}{!}{
\begin{tabular}{lcccc} 
    \toprule
    \textbf{Sequence} & \textbf{Frames} & \textbf{Area ($m^2$)} & \textbf{Total Objs} & \textbf{Dyn. Objs} \\
    \midrule
    Office-1 (Ours) & $1000$ & $9.7 \times 4.9 $ & $29$ & $3$ \\
    Bedroom-1 (Ours) & $1000$ & $5.2 \times 4.0$ & $23$ & $6$ \\
    Livingroom-1 (Ours) & $1000$ & $ 7.9 \times 5.1$ & $56$ & $11$ \\
    Kitchen-1 (Ours) & $1000$ & $6.4 \times 5.9$ & 42 & 14 \\
    Bonn\_box2~\cite{palazzolo2019iros} & $1279$ & - & $\approx 30$ & $1$ \\
    \bottomrule
\end{tabular}}
\vspace{-5mm} 
\end{table}

\vspace{2pt}
\noindent \textbf{Baselines.} 
We compare with static 3DGS mapping baselines, MonoGS~\cite{matsuki2024gaussian}, GS-ICP~SLAM~\cite{ha2024rgbd}, SplaTAM~\cite{keetha2024splatam}, and the dynamic baseline WildGS-SLAM~\cite{Zheng2025WildGS}.
To decouple mapping quality from tracking drift, \textit{all} methods utilize ground truth (GT) poses during evaluation.

\vspace{2pt}
%
%
\noindent \textbf{Metrics.}
We evaluate the system from three aspects:
\begin{enumerate}[label=\arabic*) , left=0pt, labelsep=1em]  
\item \textit{Rendering Quality:} We report PSNR, SSIM, and LPIPS, following MonoGS~\cite{matsuki2024gaussian}. Notably, all rendering metrics are computed on the dynamic Region of Interest (ROI) to avoid static-background bias.
\item \textit{Perception Accuracy:} We report 3D mAP (AP25/AP50) for object localization, following CubifyAnything~\cite{lazarow2025cubify}.
\item \textit{Mapping Efficiency:} We report average FPS to evaluate runtime efficiency.
\end{enumerate}

\vspace{2pt}
\noindent \textbf{Implementation Details.} 
Our method is implemented on the MonoGS~\cite{matsuki2024gaussian} codebase (PyTorch) and runs on an Intel i9-14900KF CPU and a single NVIDIA RTX 4090 GPU.
For alignment optimization (Sec.~\ref{sec:exp_retrieval}), we use a two-stage Adam schedule (1,000 iterations total):
a \textit{coarse stage} (first 700 iters) with learning rates $1 \times 10^{-2}$ (rot) and $5 \times 10^{-3}$ (trans), followed by a \textit{fine stage} decaying to $6 \times 10^{-3}$ and $3 \times 10^{-3}$, respectively, to ensure stable convergence.

\begin{figure*}[!t]
 \vspace{0.5em} 
\centering
	\includegraphics[width=17.5cm]{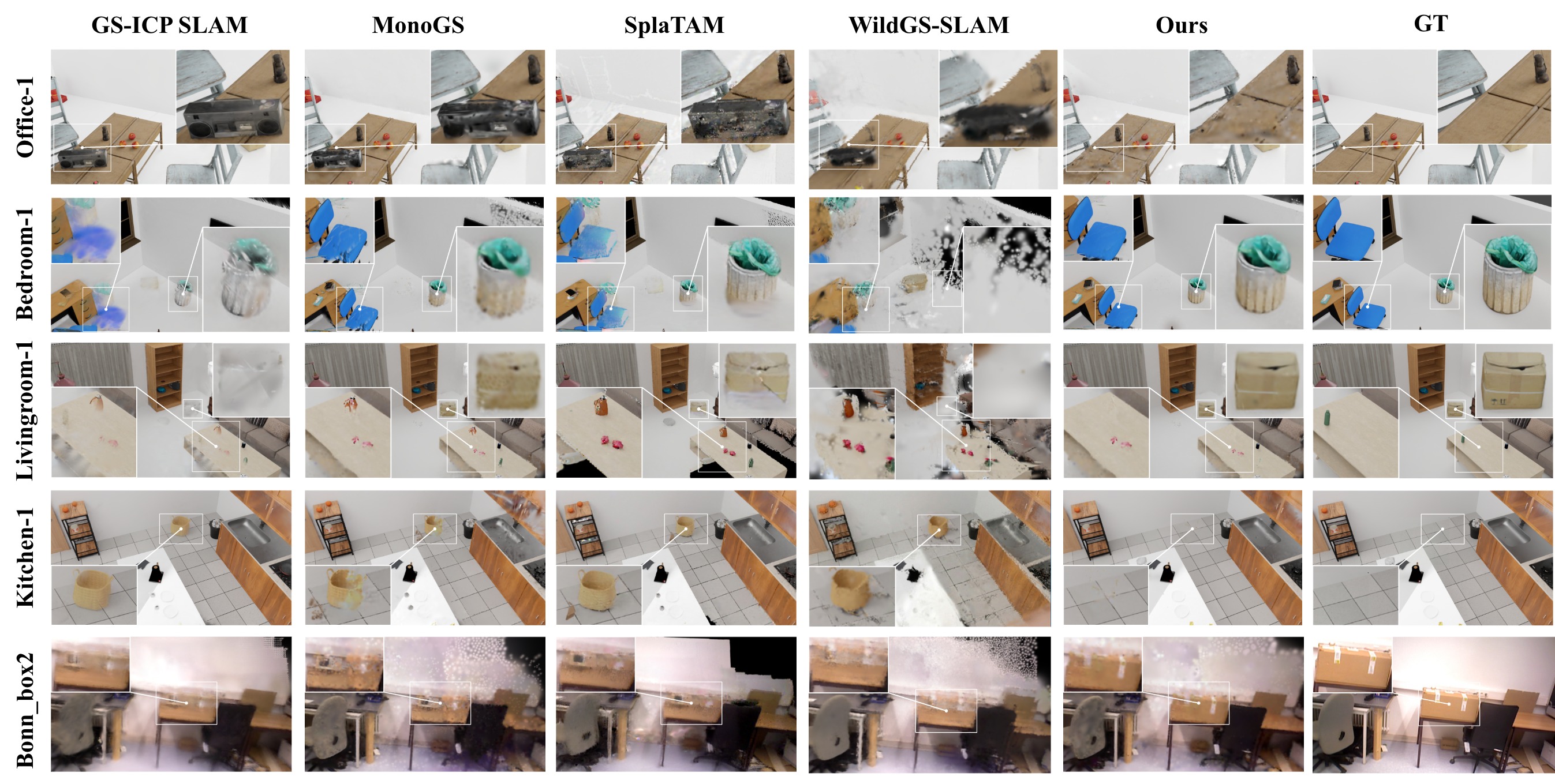}
	\caption{\textbf{Qualitative comparison of post-change reconstruction recovery.} 
        All visualizations are rendered 10 seconds after a discrete object rearrangement event. While conventional 3DGS baselines suffer from severe ghosting and blurred textures due to slow gradient-based updates, CubifyGS explicitly manages object assets to instantly eliminate artifacts and restore sharp, high-fidelity object geometry comparable to the Ground Truth (GT).}
	\label{fig:qualitative_result} 
    \vspace{-2mm}
\end{figure*}

\subsection{System-level Comparisons}
\label{sec:system_comparison}

To evaluate system-level performance, we compare CubifyGS with static and dynamic 3DGS baselines on all five benchmark sequences.
Unlike conventional full-viewpoint evaluation, we focus on quality recovery after scene changes.
For each rearrangement event (iteration 0), we evaluate PSNR/SSIM/LPIPS on dynamic ROIs every 30 optimization iterations within a fixed 10-second window.
This evaluation protocol emphasizes each method's reconstruction-repair capability in dynamic scenes.
The score reported in Table~\ref{tab:system_comparison} is obtained by first averaging these samples over time within each event and then averaging across dynamic objects.


\begin{table*}[t]
\centering
\caption{Quantitative Comparison of Dynamic Reconstruction Quality}
\label{tab:system_comparison}
\setlength{\tabcolsep}{3.5pt}
\resizebox{1.0\textwidth}{!}{
\begin{tabular}{l ccc ccc ccc ccc ccc} 
    \toprule
    \multirow{2}{*}{Method} & \multicolumn{3}{c}{Office-1} & \multicolumn{3}{c}{Bedroom-1} & \multicolumn{3}{c}{Livingroom-1} & \multicolumn{3}{c}{Kitchen-1} & \multicolumn{3}{c}{Bonn-box2} \\
    \cmidrule(lr){2-4} \cmidrule(lr){5-7} \cmidrule(lr){8-10} \cmidrule(lr){11-13} \cmidrule(lr){14-16}
     & \small{PSNR$\uparrow$} & \small{SSIM$\uparrow$} & \small{LPIPS$\downarrow$} & \small{PSNR$\uparrow$} & \small{SSIM$\uparrow$} & \small{LPIPS$\downarrow$} & \small{PSNR$\uparrow$} & \small{SSIM$\uparrow$} & \small{LPIPS$\downarrow$} & \small{PSNR$\uparrow$} & \small{SSIM$\uparrow$} & \small{LPIPS$\downarrow$} & \small{PSNR$\uparrow$} & \small{SSIM$\uparrow$} & \small{LPIPS$\downarrow$} \\
    \midrule
    MonoGS~\cite{matsuki2024gaussian} & 10.08 & 0.314 & 0.669 & 15.91 & 0.582 & 0.619 & 14.36 & 0.481 & 0.607 & 18.64 & 0.452 & 0.570 & 14.63 & 0.314 & 0.665 \\
    GS-ICP SLAM~\cite{ha2024rgbd} & 10.26 & 0.333 & 0.651 & 15.13 & 0.664 & 0.598 & 12.73 & 0.458 & 0.683 & 17.13 & 0.462 & 0.574 & 15.34 & 0.393 & 0.613 \\
    SplaTAM~\cite{keetha2024splatam} & 10.38 & 0.268 & 0.669 & 15.54 & 0.602 & 0.615 & 12.92 & 0.434 & 0.688 & 14.82 & 0.431 & 0.621 & 15.12 & 0.377 & 0.633 \\
    WildGS-SLAM~\cite{Zheng2025WildGS} & 11.61 & 0.410 & 0.677 & 13.66 & 0.543 & 0.613 & 13.09 & 0.385 & 0.727 & 12.40 & 0.415 & 0.872 & 14.11 & 0.262 & 0.695 \\
    \midrule
    \textbf{Ours} & \textbf{17.98} & \textbf{0.537} & \textbf{0.548} & \textbf{23.49} & \textbf{0.686} & \textbf{0.542} & \textbf{20.50} & \textbf{0.546} & \textbf{0.476} & \textbf{21.18} & \textbf{0.463} & \textbf{0.434} & \textbf{16.85} & \textbf{0.475} & \textbf{0.579} \\
    \bottomrule
\end{tabular}}
\vspace{-4mm}
\end{table*}

\noindent \textbf{Mapping Quality.}
As shown in Table~\ref{tab:system_comparison}, our method achieves the best overall performance across the five benchmark sequences, with an average improvement of 35.83\% in PSNR over the strongest baseline.
These results indicate that CubifyGS delivers superior reconstruction recovery under dynamic object rearrangement.
Notably, the largest margin is observed in \textit{Bedroom-1}, where our method improves PSNR by 7.58 dB compared with MonoGS\cite{matsuki2024gaussian}, highlighting its advantage in challenging maintenance cases.
To further illustrate behavior after scene changes, we provide qualitative comparisons in Fig.~\ref{fig:qualitative_result}.
Compared with baselines, CubifyGS removes ghost artifacts earlier and restores sharper object structures with fewer optimization iterations, yielding more consistent visual quality during the post-event recovery stage.

\noindent \textbf{Runtime Efficiency.} 
In addition to the reconstruction quality, we assess the runtime efficiency of our framework under identical hardware conditions. Evaluated on a single NVIDIA RTX 4090 GPU, CubifyGS achieves dynamic event detection and map maintenance at an approximate rate of \textbf{20 FPS}. In contrast, WildGS-SLAM~\cite{Zheng2025WildGS} employs continuous online training of an MLP for motion detection, resulting in a throughput of less than 0.5 FPS. Under the same setting, CubifyGS is over \textbf{40$\times$} faster than WildGS-SLAM.
    

\begin{table}[t]
\centering
\caption{Quantitative Evaluation of 3D Object Localization}
\label{tab:localization_results}
\setlength{\tabcolsep}{4pt} 
\resizebox{\columnwidth}{!}{ 
\begin{tabular}{l cc cc cc cc} 
    \toprule
    \multirow{2.5}{*}{\textbf{Scene}} & \multicolumn{2}{c}{\textbf{AP$_{25}$}} & \multicolumn{2}{c}{\textbf{AP$_{50}$}} & \multicolumn{2}{c}{\textbf{AR$_{25}$}} & \multicolumn{2}{c}{\textbf{AR$_{50}$}} \\
    \cmidrule(lr){2-3} \cmidrule(lr){4-5} \cmidrule(lr){6-7} \cmidrule(lr){8-9}
    & CA & Ours & CA & Ours & CA & Ours & CA & Ours \\
    \midrule
    Office-1 & 35.7 & 86.1 & 24.9 & 61.8 & 38.2 & 88.2 & 29.4 & 70.6 \\
    Bedroom-1 & 31.0 & 65.6 & 20.5 & 40.2 & 31.0 & 72.4 & 24.1 & 51.7 \\
    Livingroom-1 & 12.9 & 54.6 & 6.7 & 20.9 & 16.4 & 59.0 & 9.8 & 36.1 \\
    Kitchen-1 & 27.0 & 54.3 & 8.5 & 35.1 & 27.9 & 65.1 & 14.0 & 48.8 \\
    \midrule
    \textbf{Average} & \textbf{26.7} & \textbf{65.2} & \textbf{15.2} & \textbf{39.5} & \textbf{28.4} & \textbf{71.2} & \textbf{19.3} & \textbf{51.8} \\
    \bottomrule
\end{tabular}}
\vspace{-6mm}
\end{table}

\subsection{Dynamics Perception Verification}
\label{sec:exp_dynamics}

\begin{figure}[htbp]
\centering
\includegraphics[width=1.0\linewidth]{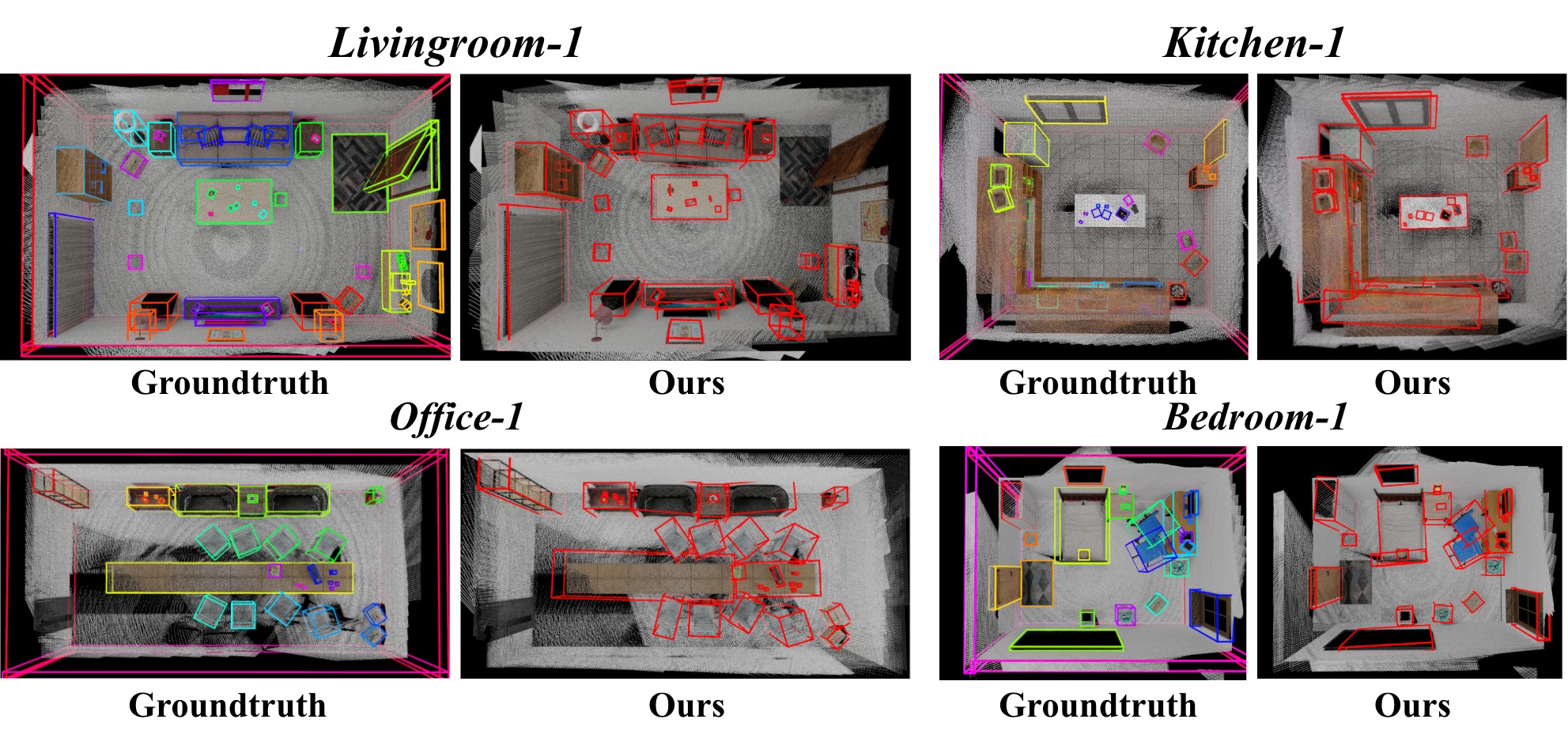} 
\caption{\textbf{Qualitative 3D localization results.} Our predicted bounding boxes (right) closely align with the ground truth (left) across representative benchmark scenes.}
\label{fig:localization_qualitative}
\end{figure}

Accurate 3D object localization is a prerequisite for stable retrieval-and-alignment initialization, directly impacting the quality of lifelong map maintenance. We evaluate this capability on our proposed benchmark using 3D Average Precision (AP) at IoU thresholds of 0.25 and 0.50, alongside Average Recall (AR), adopting the protocol from CubifyAnything~\cite{lazarow2025cubify}. To demonstrate the efficacy of our temporal design, we directly compare against single-frame CubifyAnything (CA)~\cite{lazarow2025cubify} as the baseline. The \texttt{bonn\_kidnapping\_box2} sequence is excluded from this specific evaluation due to the lack of high-fidelity 3D bounding box annotations. As shown in Table~\ref{tab:localization_results}, CubifyGS ($\text{AP}_{25}$: 65.2) substantially outperforms the single-frame CA baseline (26.7). This significant margin confirms that our multi-view association and temporal aggregation are crucial for robust perception. Furthermore, qualitative results in Fig.~\ref{fig:localization_qualitative} demonstrate tight alignment between our predicted bounding boxes and the ground truth, ensuring reliable geometric priors for downstream asset retrieval.

\subsection{Retrieval \& Alignment Analysis}
\label{sec:exp_retrieval}

\begin{figure}[htbp]
\centering
\includegraphics[width=1.0\linewidth]{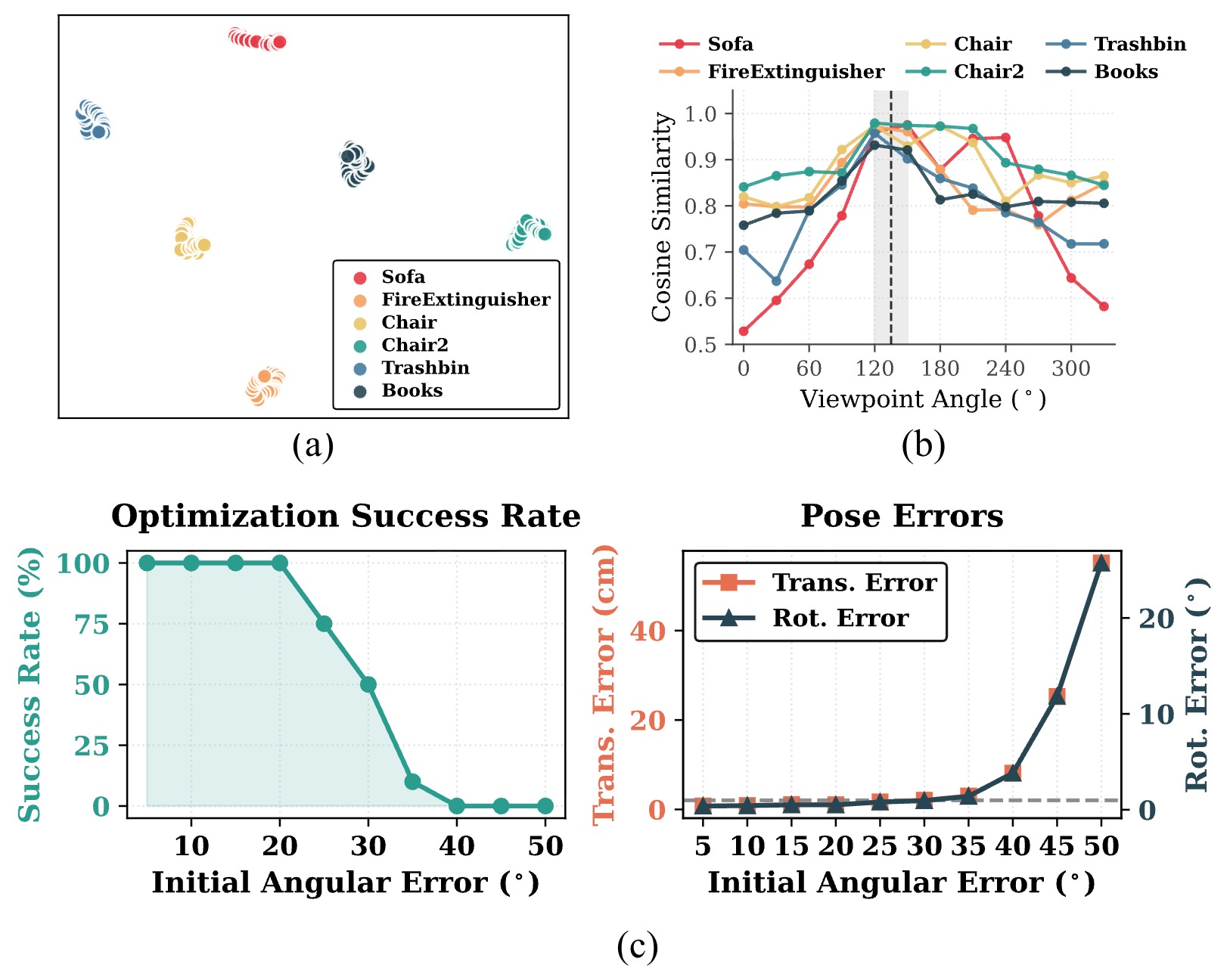} 
\caption{\textbf{Retrieval and Alignment Validation.} \textbf{(a)} t-SNE of DINOv3 features shows distinct instance clusters across varying viewpoints. \textbf{(b)} Cosine similarity consistently peaks within $\pm 15^\circ$ of the ground truth (gray band). \textbf{(c)} Fine alignment optimization succeeds for angular perturbations up to $25^\circ$ (green region), fully covering the coarse retrieval error margin.}
\label{fig:retrieval_analysis}
\vspace{-3mm}
\end{figure}



In CubifyGS, the Retrieval-and-Align module is key to rapid updates after object-level changes. Given a new observation, the system identifies object semantics, retrieves a matched canonical asset from the global library, initializes pose by coarse viewpoint matching, and then refines it with photometric fine alignment. We evaluate this module from three aspects: semantic discriminability, coarse alignment precision, and fine alignment robustness. These evaluations verify that retrieval provides reliable initialization and that fine alignment consistently recovers the remaining error.

\subsubsection{Discriminative Semantics}

To verify instance distinguishability, we extract DINOv3 features from 360-degree surround views of 6 objects ($15^\circ$ intervals). The t-SNE projection (Fig.~\ref{fig:retrieval_analysis}a) displays well-separated clusters, confirming that our semantic representation is robust to viewpoint variations and minimizes false matches.

\subsubsection{Coarse Alignment Precision}
We compute cosine similarities between a $135^\circ$ query view and reference views sampled at $30^\circ$ intervals. As plotted in Fig.~\ref{fig:retrieval_analysis}b, the maximum similarity consistently falls within $135^\circ \pm 15^\circ$ of the ground truth. This discrete retrieval successfully bounds the initial angular error to $\pm 15^\circ$.

\subsubsection{Fine Alignment Convergence}

To evaluate photometric fine alignment robustness, we initialize optimization using neighboring dataset frames separated by $5^\circ$ to $50^\circ$. We superimpose random noise ($\sigma_{rot}=2^\circ, \sigma_{trans}=1\text{cm}$) to simulate unstructured retrieval jitter and real-world perspective shifts. A trial is successful if it converges within $2\text{cm}$ and $1^\circ$ of the ground truth under 1,000 iterations. As shown in Fig.~\ref{fig:retrieval_analysis}c, our method achieves a near $100\%$ success rate for initial angular discrepancies up to $25^\circ$. This convergence basin fully encompasses the $\pm 15^\circ$ coarse retrieval error bound, guaranteeing autonomous and precise pose recovery.

\subsection{Ablation Study}

\label{sec:ablation}

\begin{table}[t]
\vspace{3mm} 
\centering
\caption{Ablation Study of Core Components on Dynamic ROI.}
\label{tab:ablation_study}
\setlength{\tabcolsep}{3.5pt} 
\resizebox{\columnwidth}{!}{
\begin{tabular}{cc ccc ccc}
\toprule
\multicolumn{2}{c}{\textbf{Component}} & \multicolumn{3}{c}{\textbf{Livingroom-1 (Ours)}} & \multicolumn{3}{c}{\textbf{Kitchen-1 (Ours)}} \\
\cmidrule(lr){1-2} \cmidrule(lr){3-5} \cmidrule(lr){6-8}
FDO & GAL & PSNR$\uparrow$ & SSIM$\uparrow$ & LPIPS$\downarrow$ & PSNR$\uparrow$ & SSIM$\uparrow$ & LPIPS$\downarrow$ \\
\midrule
\checkmark & $-$ & 15.98 & 0.502 & 0.571 & 20.11 & 0.458 & 0.488 \\
$-$ & \checkmark & 17.58 & 0.511 & 0.503 & 20.54 & 0.456 & 0.441 \\
\midrule
\checkmark & \checkmark & \textbf{20.50} & \textbf{0.546} & \textbf{0.476} & \textbf{21.18} & \textbf{0.463} & \textbf{0.434} \\
\bottomrule
\end{tabular}}
\vspace{-6mm}
\end{table}

To validate the individual contributions of our core components, we conduct an ablation study focusing on the Focus-Driven Optimization (FDO) and the Global Asset Library (GAL). We evaluate dynamic ROI rendering quality on \textit{Livingroom-1} and \textit{Kitchen-1}, which are the most challenging sequences in our benchmark according to the system-level comparison in Table~\ref{tab:system_comparison} due to high object density and complex rearrangement events. The quantitative comparisons are summarized in Table~\ref{tab:ablation_study}.

\noindent \textbf{Efficacy of Focus-Driven Optimization (FDO).} 
FDO dynamically allocates computational resources strictly to active regions undergoing physical changes. To assess its impact, we disable FDO (applying a uniform loss weight across the entire image). Without FDO, the gradient flow from the small dynamic region is overwhelmed by the vast number of converged background pixels, leading to sluggish local adaptation. As shown in Table~\ref{tab:ablation_study}, the absence of FDO results in a noticeable performance drop (e.g., PSNR decreases from 20.50 to 17.58 on \textit{Livingroom-1}), proving that active spatial weighting is crucial for rapid background inpainting and asset refinement.

\noindent \textbf{Significance of the Global Asset Library (GAL).} 
GAL provides pre-optimized object templates to enable our ``Retrieve-and-Reuse'' paradigm. To investigate its necessity, we define a ``Cold Start'' baseline where GAL is disabled. Instead of retrieving an asset, this baseline initializes new Gaussian primitives from scratch upon detecting a novel object. This implicit, gradient-based re-optimizing is highly inefficient, causing a severe PSNR drop of 4.52 dB on \textit{Livingroom-1} (from 20.50 down to 15.98). This confirms that explicitly snapping high-quality assets from GAL into the scene is essential for instant, high-fidelity recovery.

\section{CONCLUSIONS}

We propose CubifyGS, a dynamic 3DGS mapping framework for rigid object rearrangement. It shifts map maintenance from passive gradient-based re-optimization to active asset management using reusable Gaussian assets, spatio-temporal dynamics perception, and event-triggered adaptive optimization. Experiments show improved artifact suppression and maintenance efficiency over representative reproducible baselines.

However, our current study focuses on rigid object rearrangement with reusable object assets, and the promoted assets are reused as fixed templates. In future work, we will extend CubifyGS to broader dynamic scenarios, investigate incremental asset update and cross-session asset merging, improve perception robustness and efficiency, and further evaluate how rapid map maintenance benefits downstream tasks such as long-term camera localization.



\bibliographystyle{IEEEtran}
\bibliography{main.bib}

\end{document}